\documentclass[runningheads]{llncs}
\usepackage{orcidlink}
\usepackage[T1]{fontenc}
\usepackage{tabularx}
\usepackage{booktabs}       
\usepackage{multirow}
\usepackage{transparent}
\usepackage[table]{xcolor}
\usepackage{tabu}
\usepackage[T1]{fontenc}
\usepackage{cite}
\usepackage{amsmath,amssymb,amsfonts}
\usepackage{algorithmic}
\usepackage{graphicx}
\usepackage{textcomp}
\usepackage{hyperref}
\usepackage{doi}
\usepackage[capitalise]{cleveref}
\Crefname{section}{Sec.}{Secs.}
\usepackage{multirow}
\usepackage{underscore}
\usepackage{graphicx} 
\usepackage{booktabs}
\usepackage{colortbl}
\usepackage{makecell}
\usepackage{subcaption} 
\usepackage{float}

\renewcommand{\arraystretch}{1.2}

\newcommand{\cint}[2]{
  \makecell[tc]{
    #1\\[-3.2pt] 
    {\footnotesize\textcolor[gray]{0.55}{#2}}%
  }
}

\usepackage{graphicx,verbatim}

\begin{document}
%
\title{Cross-Modal Contrastive Learning of ECG and Angiography Representations for Severe Stenosis Classification}

\author{Nikola Cenikj\textsuperscript{1,4,*} \orcidlink{0009-0002-7952-786X}, Özgün Turgut\textsuperscript{1}\orcidlink{0009-0002-8704-0277}, Alexander Müller\textsuperscript{4}, Alexander Steger\textsuperscript{4}, Jan Kehrer\textsuperscript{4}, Marcus Brugger\textsuperscript{4}, Daniel Rueckert\textsuperscript{1,2,3}\orcidlink{0000-0002-5683-5889}, Eimo Martens\textsuperscript{4}\orcidlink{0000-0002-5801-0901}, and Philip Müller\textsuperscript{1}\orcidlink{0000-0001-8186-6479}}  

\authorrunning{Nikola Cenikj et al.}
\institute{\textsuperscript{1} Chair for AI in Healthcare and Medicine, Technical University of Munich and TUM University Hospital, Munich, Germany\\
\textsuperscript{2} Department of Computing, Imperial College London, UK\\
\textsuperscript{3} Munich Center for Machine Learning (MCML), Munich, Germany\\
\textsuperscript{4} Department of Internal Medicine, TUM University Hospital, Munich, Germany\\
    \email{nikola.cenikj@tum.de}
    }
  \titlerunning{Contrastive Learning of ECG and Angiography Representations for Stenosis}
\maketitle              

\begin{abstract}
Coronary artery stenosis is a common cardiovascular disease, with severe, untreated cases posing significant risks of heart attack. Although coronary (X-ray) angiograms remain the standard for stenosis diagnosis, they are invasive, time- and resource-intensive, and therefore only performed on patients with a high probability of disease based on symptoms and prior clinical tests. However, a subset of patients, especially those without symptoms, may remain undiagnosed. Detecting indications of stenosis from ECGs, which are fast, cheap, non-invasive, and thus routinely acquired even in asymptomatic patients, would support early diagnosis. However, as no reliable stenosis-specific signal has been identified in ECGs, they can not currently be used for stenosis risk stratification. To address this, we introduce \emph{StenCE}, a pretraining framework, allowing stratification of patients based on features derived directly from ECGs. Evaluations across varying stenosis severity thresholds and additional ECG disease classification tasks demonstrate consistent performance improvements across different ECG encoders, outperforming previous work. The obtained models successfully detect signals for stenosis diagnosis in ECGs and are the first to achieve high performance in severe stenosis classification. The source code
is available at \href{https://github.com/NikolaCenic/ecg-stenosis-cls}{https://github.com/NikolaCenic/ecg-stenosis-cls}. 

\end{abstract}

\keywords{Deep-Learning\and  Multi-modal Pretraining \and Coronary Artery Stenosis \and Coronary X-ray Angiography \and Electrocardiogram.}
\section{Introduction}
Coronary artery stenosis is a common cardiovascular disease that gets progressively more severe over time. As severe cases often lead to heart failure, early diagnosis is critical for improving the survival rate. Diagnosis of severe stenosis is typically performed through coronary (X-ray) angiograms, where multiple angiography views capture different segments of the coronary arteries. However, X-ray angiography is an invasive procedure with a small mortality risk and is typically reserved for patients with a high likelihood of stenosis based on symptoms and clinical tests. Consequently, asymptomatic patients may remain undiagnosed while the disease progresses. Identifying stenosis indicators from a fast, non-invasive, and routinely-acquired modality such as ECG would enable early diagnosis, even for asymptomatic patients. However, despite being used for diagnosing different cardiovascular diseases,  ECGs provide limited information for coronary stenosis, as patients with severe stenosis often have normal ECGs~\cite{ecg_limitations}.

\begin{figure}[!t]
  \centering
     \includegraphics[width=.8\linewidth]{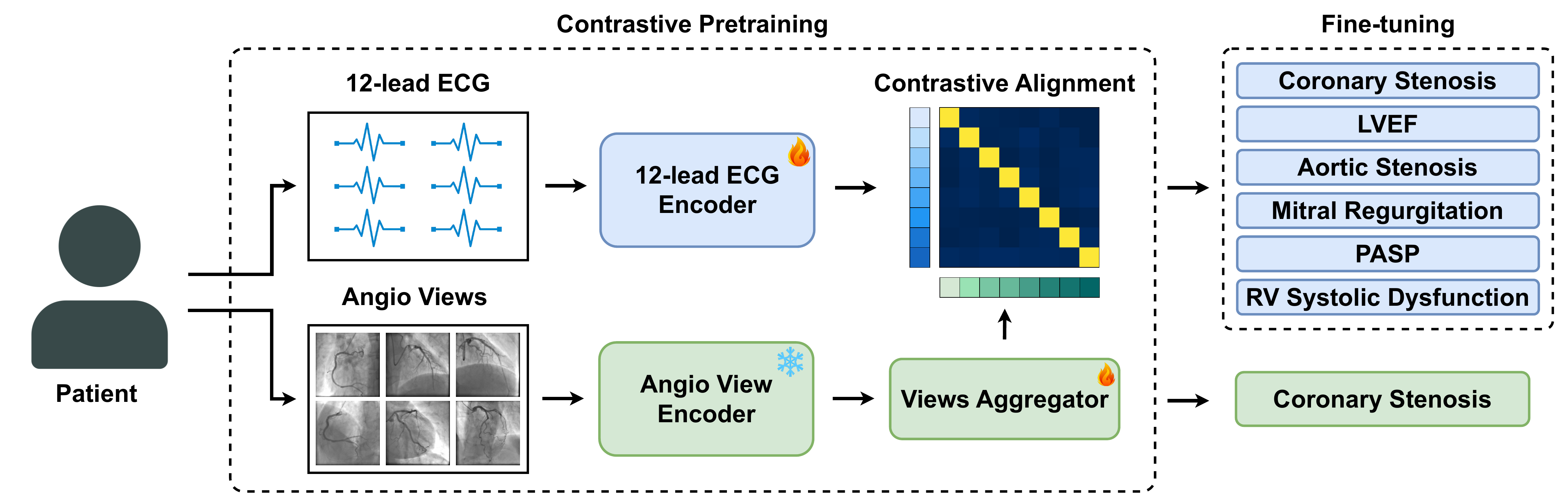}
     \caption{Overview of the proposed approach: Multi-view angiography and 12-lead ECG from the same patient are encoded with transformer-based modality-specific encoders. The ECG encoder is pretrained to extract features aligned with the frozen angiography encoder, trained for coronary stenosis classification. The pretrained ECG encoder is then fine-tuned for coronary stenosis and additional cardiac abnormalities diagnosed from ECGs.}\label{fig:main}
\end{figure}

In this work, we aim to detect signals of stenosis in ECGs and thus enable the early identification of stenosis risks. To achieve this, we develop a deep learning based stenosis classifier working only on ECG inputs. To train this model, we rely on information distillation from an X-ray angiogram encoder. More precisely, as shown in \cref{fig:main}, we employ multi-modal contrastive learning between an ECG-encoder and an angiography encoder trained for stenosis classification, followed by fine-tuning of the pretrained ECG model.
Our contributions are as follows:
\begin{enumerate}
 \item We propose \emph{StenCE}, a contrastive pretraining framework that aligns ECG representations with features from a multi-view angiography stenosis classification model, thereby enabling an ECG encoder to detect stenosis signals from ECGs alone.
 \item Our evaluation on clinical stenosis classification demonstrates an AUC of 0.822 for the most severe cases, showcasing that severe stenosis can be identified solely from ECGs, enabling early stenosis detection.
 \item Evaluations among multiple stenosis severities as well as on additional cardiac abnormalities on the  EchoNext dataset, further demonstrate the utility of our pretraining framework.
\end{enumerate}

\section{Related Work}
\noindent\textbf{Coronary Artery Stenosis.} Most existing stenosis detection work focuses on single angiographic views, without considering different views of a patient~\cite{arcade_1,arcade_2,stenosis_detection_extra_1,stenosis_detection_extra_2,stenosis_detection_overview,stenosis_detection_extra_spatio_temporal}. 
While \cite{stenosis_view_and_patient_level_eval} evaluates at patient-level, \cite{multi_view_stenosis} is the only study training on the patient-level, using all available views. ECGs, however, have been part of general time-series pretraining~\cite{otis} as well as ECG-specific pretraining~\cite{xecg,ecgfounder}, but coronary stenosis diagnosis from ECGs remains challenging~\cite{ecg_limitations}.
In~\cite{screening_stenosis_from_ecg_plus_tests}, 12-lead ECGs are used to classify stenosis ($\le 50\;|\ge99$), yielding an AUC of 0.654. Performance increases to 0.717 when using only clinical risk factors and further to 0.847 when combining the two modalities.
The work in~\cite{stenosis_from_ecg} trains a model for ECG stenosis classification ($<70\;|\ge70$), and achieves an AUC of 0.57. Both \cite{screening_stenosis_from_ecg_plus_tests} and \cite{stenosis_from_ecg} show that ECGs alone are not sufficient for reliable stenosis diagnosis. However, neither of them leverages cross-modal pretraining for the extraction of stenosis-relevant features out of the ECG signal.

\noindent\textbf{Cross-modal knowledge transfer.} 
Cross-modal knowledge transfer closes the gap between two modalities by aligning their feature representations. Initially designed for contrastive language-image pairing (CLIP)~\cite{clip}, it has been widely adopted in the medical domain, for tasks including radiology report generation\cite{clip_radiology_and_text}, anomaly detection~\cite{clip_anomaly_detection}, and biomedical foundation models~\cite{biomedical_clip}. In cardiology, cross-modal pretraining has been used to align ECG representations with modalities such as echocardiograms~\cite{echoing_ecg} and cardiac MRIs~\cite{ozgun_cmr,ptacl}. However, no prior work has explored contrastive pretraining between angiography and ECG pairs, despite them being suitable for cross-modal alignment.

\begin{table}[t!]

\setlength{\tabcolsep}{0.2em}
\caption{Distribution of patients, number of pairs, and stenosis samples based on different severity thresholds across train, validation, and test splits.}\label{Table:data_balance}
\centering
\footnotesize
\renewcommand{\arraystretch}{0.9}
\begin{tabular}{ccccccccc}
\toprule
&&&\multicolumn{6}{c}{\textbf{Samples per Severity Category}}\\
\cmidrule{4-9}
\textbf{Split}&\textbf{Patients}&\textbf{Pairs}&=0|=100 & =0|$\ge$90&=0|$\ge$70&$\le$50|$\ge$99&$\le$25|$\ge$70&<70|$\ge$70\\
\midrule

Train&3001&3587&   $762|533\phantom{00}$  & $762|1338$     & $762|1699$&$1888|856$&$1427|1699$& $1888|1699$  \\
Val&204&234&  $41|27$\phantom{00}   &  $41|81\phantom{0}$&$41|112$&$122|46$&$\phantom{0}83|112$&  $122|112$\\
Test&209&246&$40|32\phantom{00}$  &    $40|98\phantom{0}$  &$40|127$&$119|57$&$\phantom{0}81|127$&  $119|127$  \\

\bottomrule

\end{tabular}

\end{table}
\section{Materials and Methods}
\subsection{Datasets}\label{sec:data}

\noindent\textbf{ECG-Angio Stenosis Dataset.} We use a hospital dataset of 3,414 patients (70\% male, with a mean BMI of 27), who underwent a X-ray angiogram and have a paired 10-second 12-lead ECGs, captured up to 2 days prior to the angiogram. As some patients have multiple such ECGs, the total number of matched pairs is 4,067. For each patient, we have annotations
for the blockage in the segments of the coronary arteries~\cite{sintax_score}, extracted from structured reports done by clinicians, used for clinical decision making. The blockage annotations are categorized into: 0\%, 25\%, 50\%, 70\%, 90\%, 99\%, and 100\%, 
and a patient-level label is obtained as the maximum segment-level severity. We split the patients into disjoint train, validation, and test sets, with the distribution within each split shown in ~\cref{Table:data_balance}.

\noindent\textbf{EchoNext Dataset.} To further assess the ECG encoders, we also evaluate them on EchoNext~\cite{echonext}, a public dataset of ECGs with labels for different cardiac abnormalities derived from echocardiograms. We evaluate on the following binary tasks: left ventricular ejection fraction, pulmonary artery systolic pressure, moderate aortic stenosis, moderate mitral regurgitation, and moderate right ventricular systolic dysfunction.\label{sec:echonext}

\begin{figure}[!t]
  \centering
     \includegraphics[width=.8\linewidth]{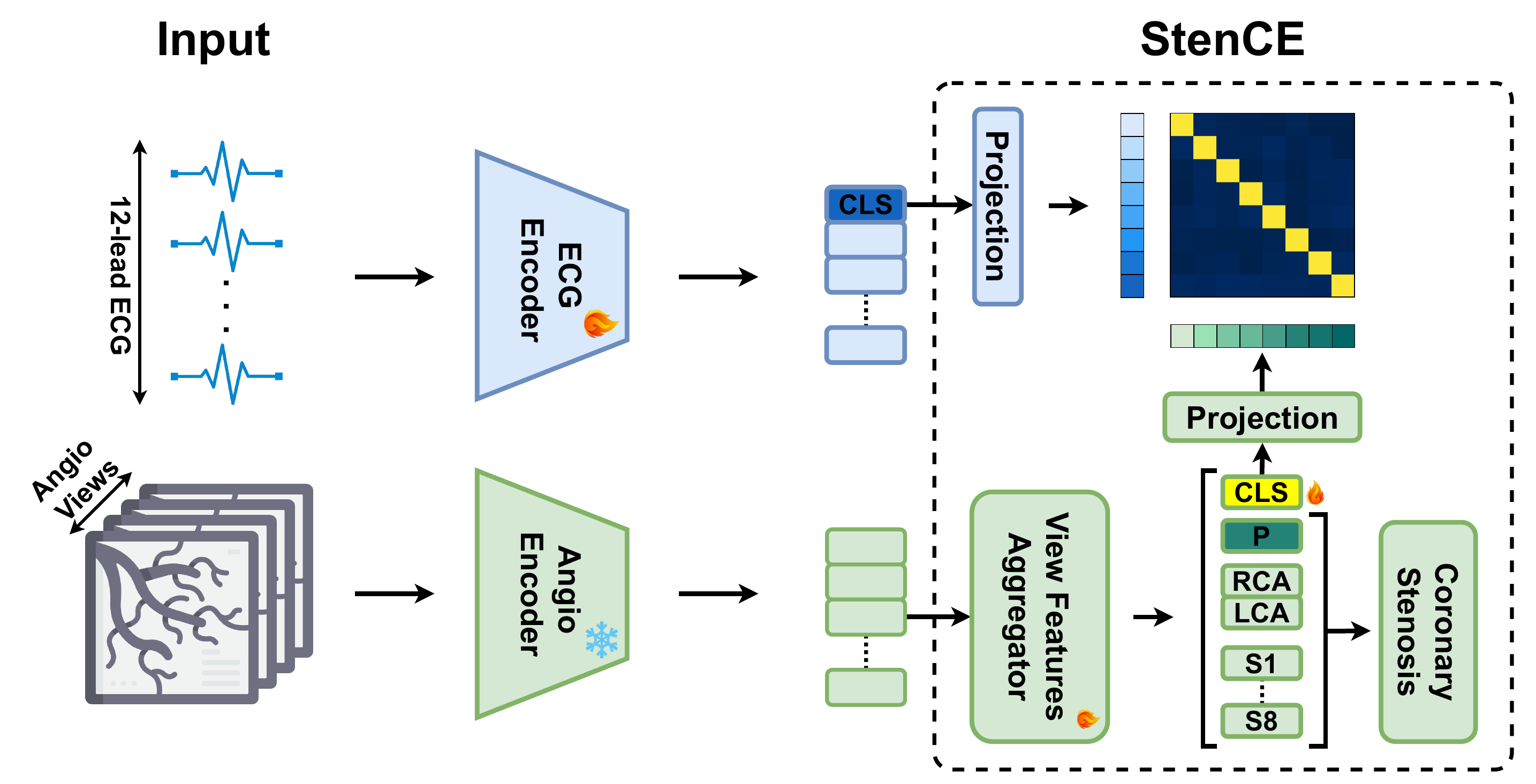}
     \caption{Given paired ECG and coronary angiography data from a single patient, both modalities are encoded and projected into a shared embedding space using transformer-based encoders and linear projections. Using contrastive loss, the ECG encoder learns to extract stenosis-relevant features, which are then utilized in fine-tuning for detecting severe stenosis using only a 12-lead ECG signal.  }\label{fig:method}
\end{figure}
\subsection{Method}

We propose \emph{StenCE}, a stenosis-driven contrastive pretraining framework for a 12-lead ECG encoder. Given a 12-lead ECG paired with coronary angiography data for patient $i$, each modality is processed with a dedicated encoder. During pretraining, the ECG encoder is trained to extract features aligned with those produced by the angiography encoder using a contrastive loss, after which the ECG encoder is fine-tuned for downstream tasks, including stenosis classification.

\noindent\textbf{ECG Encoders:}
We explore two different options for the ECG encoder: (i) \emph{OTIS} \cite{otis}, a $7.1$M-parameter transformer pretrained on time-series data including ECGs, i.e. without any cross-modal knowledge transfer, and (ii) \emph{EchoingECG} \cite{echoing_ecg}, a $128M$-parameter 1D-ResNet pretrained on triplets of echocardiogram-text-ECG, i.e. with knowledge transfer from echocardiograms. 

\noindent\textbf{Angiogram Encoder:}
The angiogram encoder is based on the SegmentMIL~\cite{multi_view_stenosis} model. It consists of an angio view encoder, encoding individual views of coronary X-ray angiogram studies, and a views aggregator component to compute study-level predictions. The view aggregator is based on a transformer decoder with query tokens for study-level, artery-level, and segment-level stenosis predictions. It was pretrained jointly on all these three tasks. 

\noindent\textbf{Knowledge Transfer through Contrastive Alignment:}
To transfer the knowledge from the angiography encoder to the ECG encoder, we follow the approach in \cite{ozgun_cmr} on pairs of individual ECGs and the associated angiography studies. More precisely, we apply the CLIP loss \cite{clip} on pairs of ECG and angiography features projected into a shared space using linear projections. To compute study-level angiography features, we introduce a new \texttt{CLS} query token into the views aggregator (extending the existing stenosis query tokens) and take its output features. We keep the angio view encoder frozen, while training the views aggregator, the ECG encoder, and the projections. 

\noindent\textbf{Supplementary Stenosis Loss:}
During contrastive alignment, we also keep training study-level, artery-level, and segment-level stenosis classification on the respective query tokens of the angiography views aggregator, following the procedure from the original SegmentMIL~\cite{multi_view_stenosis}. We balance this loss with the CLIP loss using the coefficient $W_\text{Sten} = 0.3$.

\noindent\textbf{Fine-Tuning:} After pretraining, we only keep the ECG encoder and fine-tune it on stenosis classification. Here we use only the ECG samples but derive their stenosis targets from their associated angiography studies. To address label imbalance, we use the class-weighted binary cross-entropy loss.

\subsection{Experiments}
\noindent\textbf{Baselines.}
We group the baselines into single- and cross-modal pretrained ECG encoders. As single-modality approaches, we consider \emph{OTIS}~\cite{otis}, \emph{xECG}~\cite{xecg},  and \emph{ECGFounder}~\cite{ecgfounder}. 
OTIS is pretrained on general time-series data, including ECGs, while xECG and ECGFounder are specifically pretrained using ECGs. As cross-modal pretrained models, we include \emph{EchoingECG}~\cite{echoing_ecg} and \emph{PTACL}~\cite{ptacl}, both pretrained using contrastive learning. EchoingECG uses pairs of echocardiograms, text and ECGs, while PTACL focuses on pairs of ECGs and CMRs. 


\noindent\textbf{Evaluation Setup.} 
We evaluate the learned ECG encoders using linear probing and full fine-tuning on stenosis classification across multiple severity thresholds. Additionally, we evaluate the encoders on EchoNext's tasks, formulated as multi-label classification. 
 We report the mean AUC and the 95\% confidence interval computed over three independent random seeds. We analyze the statistical significance of the performance differences using Welch's t-test ($p< 0.05$).

\begin{table}[t!]

\setlength{\tabcolsep}{0.35em}
\caption{Comparison of OTIS-StenCE and EchoingECG-StenCE encoders against OTIS, EchoingECG, and other ECG baselines under both full fine-tuning and linear probing. We report mean AUC with 95\% confidence intervals over three seeds for stenosis classification at different severity thresholds, and mean AUC across five EchoNext tasks. The best results for OTIS and EchoingECG encoders are \textbf{bold}. The results indicate that our models have strong performance for stenosis classification, particularly at the highest severity threshold ($=0\;|=100$). Furthermore, StenCE pretraining consistently boosts performance for both OTIS and EchoingECG across linear probing and full fine-tuning.}\label{Table:stenosis}
\centering
\footnotesize
\begin{subtable}{\textwidth}
\begin{tabular}{ccccc}

\toprule
&&\multicolumn{2}{c}{\textbf{Stenosis AUCs}}&\textbf{EchoNext}\\
\cmidrule(lr){3-4}
\cmidrule(lr){5-5}
\textbf{Model}&\textbf{Training}&$=0\;|=100$&$=0\;|\ge90$&Mean AUC\\
\midrule
xECG&\multirow{3}{*}{ECG}&\cint{0.672}{0.664, 0.683}&\cint{0.646}{0.625, 0.663}&\cint{0.774}{0.772, 0.777}\\
ECGFounder&&\cint{0.623}{0.588, 0.673}&\cint{0.682}{0.661, 0.693}&\cint{0.771}{0.768, 0.773}\\

PTACL&ECG+CMR&\cint{0.530}{0.520, 0.536}&\cint{0.489}{0.486, 0.492}&\cint{0.766}{0.763, 0.769}\\
\midrule
EchoingECG&ECG+Echo+Text&\cint{0.719}{0.705, 0.734}&\cint{\textbf{0.704}}{0.696, 0.717}&\cint{0.765}{0.760, 0.768}\\
    \rowcolor{cyan!15}
\rowcolor{cyan!15}
EchoingECG - StenCE&ECG+Angio& \cint{\textbf{0.822}}{0.817, 0.828} & \cint{0.668}{0.640, 0.696}& \cint{\textbf{0.770}}{0.764, 0.774}\\

\midrule

OTIS&ECG&\cint{0.650}{0.624, 0.672}&\cint{0.661}{0.651, 0.673}&\cint{0.780}{0.778, 0.782}\\
\rowcolor{cyan!15}
\rowcolor{cyan!15}
OTIS - StenCE&ECG+Angio&\cint{\textbf{0.692}}{0.678, 0.702\phantom{ }}& \cint{\textbf{0.667}}{0.650, 0.680}&\cint{\textbf{0.783}}{0.780, 0.784}\\
\bottomrule
\end{tabular}
   \caption{Fine-tuning Performance}\label{tab:main_ff}
\end{subtable}
\setlength{\tabcolsep}{0.3em}
\begin{subtable}{\textwidth}
\begin{tabular}{ccccc}
\toprule
&&\multicolumn{2}{c}{\textbf{Stenosis AUCs}}&\textbf{EchoNext}\\
\cmidrule(lr){3-4}
\cmidrule(lr){5-5}
\textbf{Model}&\textbf{Training}&$=0\;|=100$&$=0\;|\ge90$&Mean AUC\\
\midrule

EchoingECG&ECG+Echo+Text&\cint{0.643}{0.636, 0.646}&  \cint{0.625}{0.609, 0.634}&\cint{0.667}{0.662, 0.67}\\

\rowcolor{cyan!15}

EchoingECG - StenCE&ECG+Angio&  \cint{\textbf{0.816}}{0.806, 0.829}  &  \cint{\textbf{0.637}}{0.636, 0.638} & \cint{\textbf{0.739}}{0.738, 0.740}   \\
\midrule

OTIS&ECG&\cint{0.592}{0.585, 0.599}&\cint{0.503}{0.493, 0.510}&\cint{0.678}{0.672, 0.683}\\
\rowcolor{cyan!15}

OTIS - StenCE&ECG+Angio&\cint{\textbf{0.676}}{0.674, 0.677\phantom{ }}&\cint{\textbf{0.625}}{0.624, 0.628\phantom{ }}& \cint{\textbf{0.695}}{0.685, 0.710\phantom{ }}\\
\bottomrule
\end{tabular}
\caption{Linear probing Performance}
        \label{tab:main_lin_probe}
\end{subtable}

\end{table}
\section{Results}
In \cref{Table:stenosis}, we report the results under both full fine-tuning and linear probing for EchoingECG and OTIS backbones, pretrained with StenCE, referred to as EchoingECG-StenCE and OTIS-StenCE. We compare them against OTIS, EchoingECG, as well as the other ECG baselines.

\noindent\textbf{Our models are capable of detecting clear diagnostic signals of severe stenosis from ECGs alone.} The full fine-tuning results in \cref{tab:main_ff} demonstrate strong performance at the most severe stenosis threshold, achieving an AUC of 0.822 for the ($=0\;|=100$) setting. Our model is the first to achieve such high performance on ECG-based stenosis classification, confirming its ability to extract a strong diagnostic signal directly from ECGs, thereby supporting early severe stenosis detection.

\noindent\textbf{StenCE pretraining yields performance improvement in the majority of tasks for both OTIS and EchoingECG.} In full fine-tuning setting, OTIS-StenCE significantly outperforms OTIS by 4\% AUC on the severity threshold ($=0\;|=100$) , and insignificantly on ($=0\;|\ge90$). For EchoingECG, fine-tuning EchoingECG-StenCE leads to an 11\% AUC improvement on the ($=0\;|=100$) threshold, achieving the best overall performance among all models. However, on the ($=0\;|\ge90$) threshold, EchoingECG outperforms EchoingECG-StenCE by 4\% AUC. On the EchoNext tasks, all models achieve comparable fine-tuning results due to the large scale of the EchoNext dataset. Nevertheless, for both OTIS and EchoingECG, the StenCE variants retain an advantage over their base versions. The benefit of StenCE pretraining is even more pronounced in the linear probing setting (\cref{tab:main_lin_probe}), where for both OTIS and EchoingECG, the StenCE variants significantly outperform the base models on each task. The largest gain is observed for EchoingECG-StenCE, with a 17\% AUC improvement over EchoingECG on the ($=0\;|=100$) threshold. Similarly, OTIS-StenCE improves over OTIS by 8\% and 12\% AUC on the ($=0\;|=100$) and ($=0\;|\ge90$) thresholds, respectively. Similarly, on EchoNext, the StenCE models in linear probing, outperform OTIS by 2\% and EchoingECG by 7\% AUC.

\begin{figure}[!t]
  \centering
     \includegraphics[width=.9\linewidth]{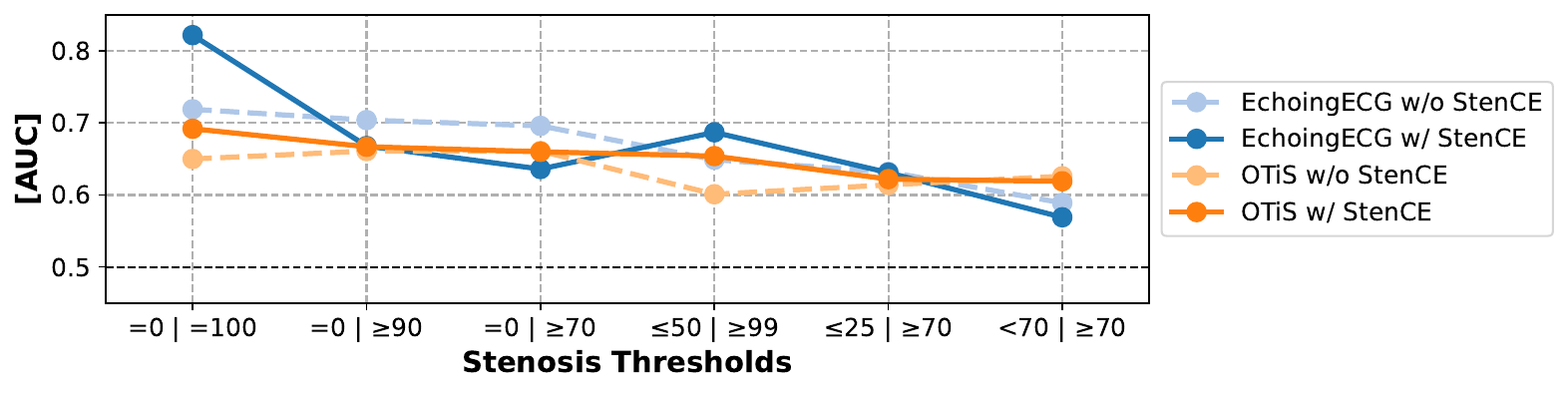}
     \caption{Comparison of the fine-tuned performance on different stenosis thresholds. While, as expected, the detection performance drops when trying to identify less severe cases, our StenCE framework still enables the separation of severe cases from both healthy (0\% blockage) and mild (50\% blockage) cases.}\label{fig:thresholds}
\end{figure}

\noindent\textbf{Detection performance drops for less severe cases.}
In \cref{fig:thresholds}, we compare the fine-tuned performance on different stenosis severity thresholds. 
Starting from an AUC of 0.822 for the most severe cases, performance decreases to 0.704 for the less severe ($=0\;|\ge90$) threshold, and approaches random performance for the hardest to diagnose stenosis with severity of  ($<70\;|\ge70$). The performance at this threshold indicates that the model is not yet suitable for clinical use. This trend is consistent with prior findings in \cite{screening_stenosis_from_ecg_plus_tests} and \cite{stenosis_from_ecg}. Although their models and datasets are not publicly available, which prevents direct comparison, our results indicate consistent improvements. In \cite{stenosis_from_ecg}, an AUC of 0.57 is reported for the ($<70\;|\ge70$) threshold, whereas our model surpasses this by more than 5\%. Similarly, in \cite{screening_stenosis_from_ecg_plus_tests}, the ECG-only model achieves an AUC of 0.654 for the ($\le50\;|\ge99$) threshold, while our model attains 0.687 on the same threshold using our dataset.

\noindent\textbf{Introducing stenosis supervision during StenCE pretraining improves performance.}
\cref{Table:ablation} shows an ablation analyzing the impact of architectural choices and the stenosis supervision in StenCE. We evaluate using OTIS and EchoingECG as ECG encoder backbones, assessing performance on stenosis classification at the ($=0\;|=100$) severity threshold and on EchoNext. For architecture, we consider a setting in which the angiography encoder is frozen and, instead of introducing a \texttt{CLS} token, we use the study-level representation.
To examine the effect of stenosis supervision, we vary the weight $W_{\text{Sten}}$ (0, 0.3, and 1). Overall, incorporating stenosis supervision with a moderate weight ($W_{\text{Sten}} = 0.3$), while still keeping the pretraining primarily driven by the CLIP loss, consistently improves performance. This configuration achieves the best results in three out of four evaluations, and only ranks second, with a 1\% difference, in stenosis classification using OTIS. 

\begin{table}[t!]
\setlength{\tabcolsep}{0.2em}
\caption{Ablation study on architectural design and stenosis supervision in StenCE. We investigate using a frozen angiography encoder and no \texttt{CLS} token, as well as different values of $W_{\text{Sten}}$. Experiments are conducted in linear probing using OTIS- and EchoingECG-based ECG encoders. We report the AUC for severe stenosis and the EchoNext tasks. The best performing model in each category is \textbf{bold}.
Results show that unfrozen angiography encoder with a \texttt{CLS} token with a moderate stenosis supervision ($W_{\text{Sten}}$ = 0.3) yields the best performance. }\label{Table:ablation}
\centering
\footnotesize
\begin{tabular}{ccccc}
\toprule
&\multicolumn{2}{c}{\textbf{OTIS}}&\multicolumn{2}{c}{\textbf{EchoingECG}}\\
\cmidrule(lr){2-3} \cmidrule(lr){4-5}

\textbf{Ablation}&$=0\;|=100$&EchoNext&$=0\;|=100$&EchoNext\\
\midrule
Frozen Angio Encoder \& no \texttt{CLS} token&\textbf{0.686}&0.691&\underline{0.766}&0.727\\
W$_{\text{Sten}}$ = 0\phantom{.3}&0.665&\underline{0.692}&0.725&0.736\\
\rowcolor{cyan!15}
W$_{\text{Sten}}$ = 0.3&\underline{0.676}&\textbf{0.695}&\textbf{0.816}&\textbf{0.739}\\
W$_{\text{Sten}}$ = 1\phantom{.3}&0.659&0.686&0.706&\underline{0.738}\\

\bottomrule

\end{tabular}

\end{table}
\section{Discussion and Conclusion}

\noindent\textbf{Limitations.}
This study has three main limitations. First, each patient in our dataset underwent an X-ray angiogram, performed due to stenosis symptoms. While this is a selection bias that should be controlled for, such clinical studies might require applying X-ray angiograms on healthy patients, exposing them to severe risks. Second, our models rely only on the ECG signal, and unlike prior work that uses clinical risk factors~\cite{screening_stenosis_from_ecg_plus_tests}, we do not integrate complementary data sources. Incorporating such data may further improve the performance.

\noindent\textbf{Conclusion.} Current models for stenosis classification rely on using backbones pretrained on ECG-specific tasks, limiting their ability to capture stenosis-specific information. We overcome this by introducing cross-model pretraining with paired ECG–angiography data, enabling the ECG encoder to encode stenosis-relevant features seen in angiography. Although the obtained performances are still not sufficient for usage in clinical practice, our results demonstrate the benefit of cross-modal pretraining and enable future developments towards preliminary assessment of coronary artery stenosis, supporting early diagnosis.

\subsubsection{Acknowledgments.} This study was approved by the Ethics Committee of TUM Klinikum Rechts der Isar (reference number 2025-395-S-CB, application dated July 13, 2025).

\bibliographystyle{splncs04}

\bibliography{sections/references}
\end{document}